\documentclass[10pt,twocolumn,letterpaper]{article}

\usepackage{cvpr}              
\definecolor{cvprblue}{rgb}{0.21,0.49,0.74}
\usepackage[pagebackref,breaklinks,colorlinks,allcolors=cvprblue]{hyperref}
\usepackage{capt-of}
\usepackage{multirow}
\usepackage{booktabs, siunitx, makecell}
\usepackage{colortbl}
\newlist{mylist}{itemize}{1}
\setlist[mylist]{label=\textbullet}

\def\paperID{37793}
\def\confName{CVPR}
\def\confYear{2026}

\title{KASALv2: Fully Automatic 3D Rotational Symmetry Classification and Axis Localization}

\author{
	Mengxin Zhang\thanks{Co-first authors (equal contribution).}
	\quad
	Yulin Wang\footnotemark[1]
	\quad
	Chen Luo\thanks{Co-corresponding authors.}
	\quad
	Yongzhe Li
	\quad
	Yijun Zhou\footnotemark[2]\\
	School of Mechanical Engineering, Southeast University\\
	Nanjing, China\\
	{\tt\small \{mx.zhang, yulinwang, chenluo, yongzheli, zhouyj\}@seu.edu.cn}
}

\begin{document}
\maketitle
\begin{abstract}
Rotational symmetry is an important prior in 6D pose estimation, improving pose accuracy and supporting symmetry-aware evaluation. However, current symmetry annotations for 3D objects remain largely manual or semi-automatic, often requiring predefined types or orders, which limits scalability. This work introduces a fully automatic, reference-free framework for symmetry-type classification, rotational-order identification, and full-axis localization across all eight canonical 3D rotational symmetry types. The method localizes a dominant high-order axis, infers its rotational order through self-consistency analysis, and reconstructs the complete symmetry structure under a hierarchy-guided formulation. A texture-aware extension further models appearance-induced reductions in rotational order while preserving axis orientations. Experiments on idealized and real-world datasets demonstrate strong accuracy and generalization, achieving 94.75\% accuracy on 438 symmetric objects in GSO. Training FoundationPose with these priors improves accuracy by up to 0.9\% across five BOP datasets, showing that automatically estimated rotational priors improve downstream 6D pose estimation. Code is available at \url{https://github.com/WangYuLin-SEU/KASAL}.
\end{abstract}    
\section{Introduction}
\label{sec:1_Introduction}
Rotational symmetry is a key geometric prior in 6D pose estimation\cite{bib1,bib3,bib4,bib5,bib6,bib7,bib2}. It resolves pose ambiguity, stabilizes optimization, and supports symmetry-aware metrics such as ADD-S\cite{ADDS}, MSSD, and MSPD\cite{MSSD}. Recent systems showing HccePose(BF)\cite{hcce} and ZebraPose\cite{bib10} use symmetry priors to improve accuracy, indicating that reliable symmetry information benefits high-precision estimation. Meanwhile, the field is moving toward unseen-object generalization and large-scale synthetic training. These trends rely on datasets with tens of thousands of 3D models, where per-object rotational symmetry is needed but manual annotation is infeasible. Such scalable symmetry annotation also benefits mesh generation and large-scale 3D asset processing.
\begin{figure}[t]
	\centering
	\includegraphics[width=0.85\linewidth]{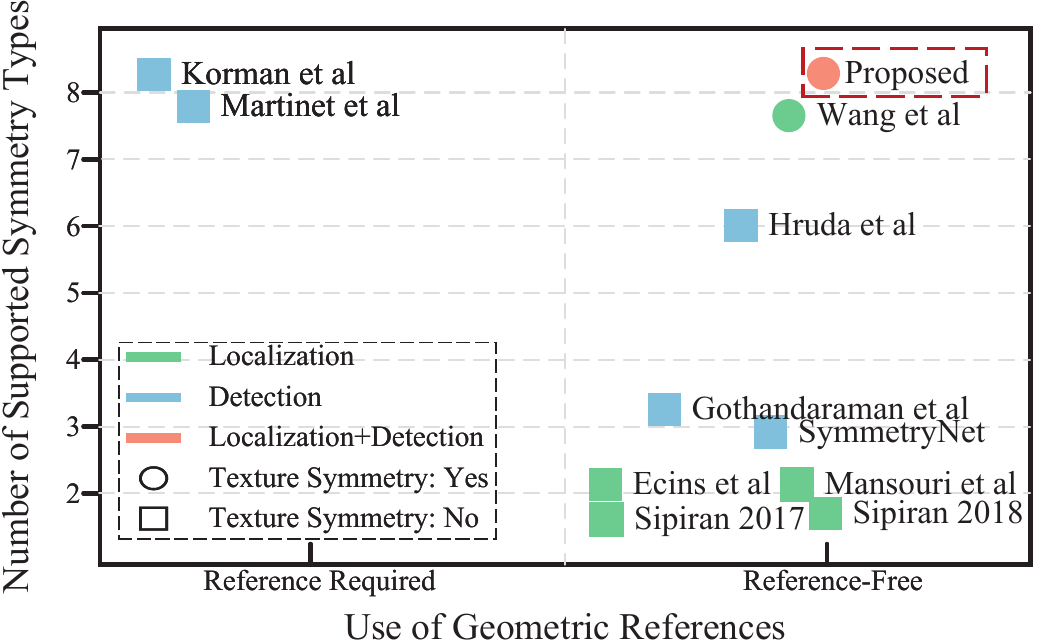}
	\caption{Comparative landscape of rotational symmetry analysis methods categorized by geometric reference dependency and task capability. The proposed method supports all eight canonical types without geometric references and handles texture-induced symmetries.
	}
	\label{fig1:mothedlocation}
\end{figure}

In addition, rotational symmetry is structurally diverse, spanning eight canonical forms\cite{geo3,geo4,geo5,geo6,geo7} with distinct axis multiplicities, rotational orders, and geometric configurations. This diversity produces many subtypes and requires a unified framework capable of joint reasoning over type and axis structure. However, as shown in Fig.~\ref{fig1:mothedlocation}, existing methods remain limited in automation and coverage: many address only subsets of symmetry types,\cite{sym-bib3,sym-bib4,sym-bib5,sym-bib6,sym-bib7,sym-bib8,sym-bib9} others depend on geometric references,\cite{sym-bib1,sym-bib2} and recent all-type approaches\cite{kasal} still require manual specification of rotational orders or categories. For large-scale unseen-object pose estimation, an automatic system capable of full symmetry classification and axis localization is needed.

To address this requirement, we extend the key-axis-based localization method of Wang et al.~\cite{kasal}, which relies on manually specified symmetry types and orders, and present KASALv2, a unified framework for complete rotational-symmetry inference across all eight canonical symmetry types without any prior input or predefined geometric references. KASALv2 transforms the task from type-conditioned axis localization into fully automatic analysis that jointly determines the symmetry type, rotational order, and the full set of canonical axes. Beyond removing manual input, it replaces the exhaustive directional enumeration of \cite{kasal} with the Geometric Degeneration Hierarchy (GDH), a computable reasoning mechanism that encodes type-specific geometric relations and guides the symmetry analysis. Guided by GDH, the system localizes a dominant high-order axis, estimates its rotational order through periodic self-consistency, infers the corresponding symmetry family, and reconstructs all canonical axes in a unified manner, with a texture-aware extension that accounts for appearance-induced reductions in rotational order.

Extensive experiments on DSRSTO\cite{kasal}, BOP\cite{VSD}, and GSO\cite{gso} validate generalization and scalability, achieving 94.75\% accuracy on real-world symmetric objects and demonstrating robust performance on the DSRSTO texture subset under appearance-induced degradation. When used as symmetry priors for FoundationPose,\cite{FP} the method improves BOP accuracy by up to 0.9\%, confirming the value of precise symmetry estimation for downstream pose prediction.
\begin{figure*}[t]
	\centering
	\includegraphics[width=1.0\linewidth]{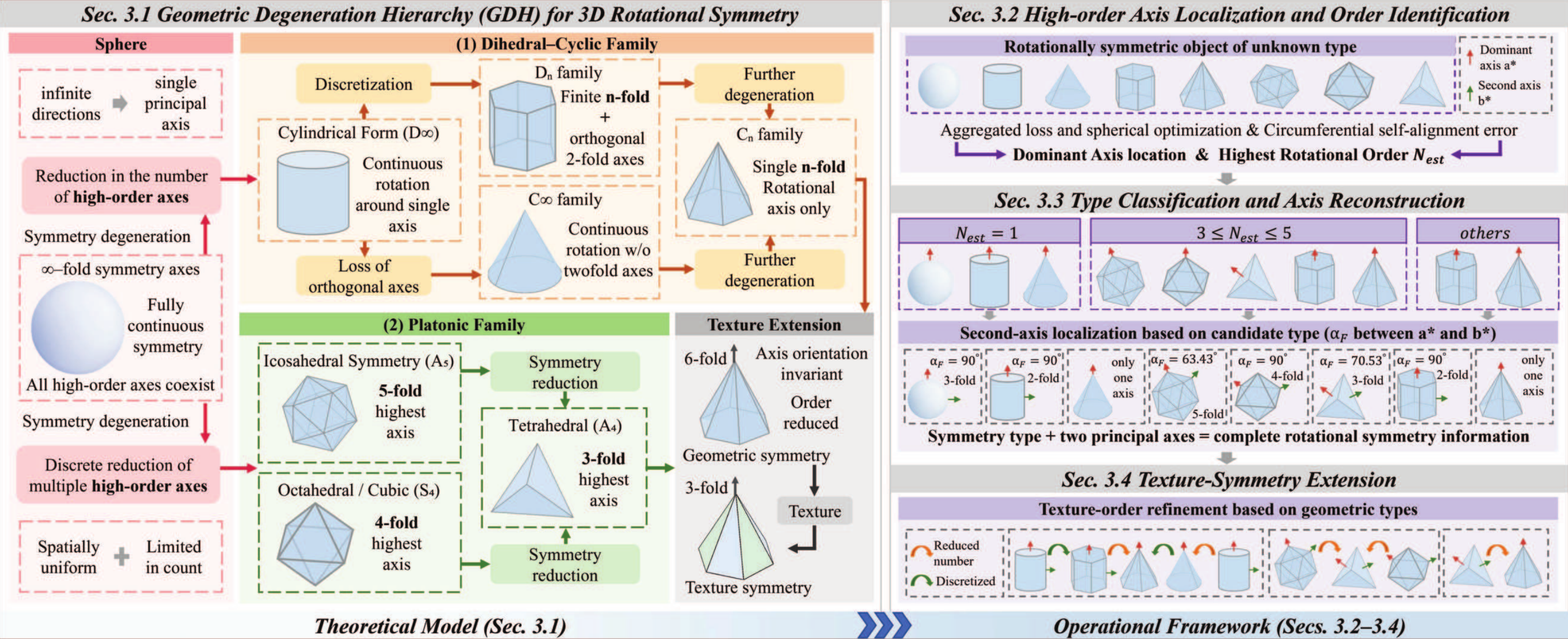}
	\caption{The pipeline integrates a Geometric Degeneration Hierarchy (GDH) and an operational framework. The GDH establishes the hierarchical taxonomy of 3D rotational symmetry, while the operational framework implements it. Given a 3D object of unknown symmetry, the system first localizes a dominant high-order axis and infers its rotational order through self-consistency optimization. Based on this order, candidate symmetry types are derived and verified by checking whether a secondary axis satisfies the corresponding geometric constraints. This hierarchical verification yields type classification and reconstruction of all canonical axes under the identified geometric relations. Finally, the texture-symmetry module refines the classification by modeling appearance-induced degradation, covering reduced, invariant, and discretized texture cases.
	}
	\label{fig2:pipline}
\end{figure*}
Our contributions are summarized as follows:
\begin{mylist}[itemsep=3pt,topsep=1pt,parsep=0pt,partopsep=0pt]
	\item A unified and fully automatic framework for 3D rotational symmetry analysis that performs symmetry-type classification, rotational-order identification, and full-axis localization across all eight canonical types in a reference-free pipeline.
	\item A geometry-driven formulation, the Geometric Degeneration Hierarchy (GDH), introduced as a computable geometric hierarchy that organizes rotational symmetries via structured axis reduction and provides the algorithmic pathway for mapping an unknown object to its canonical symmetry family.
	\item An operational instantiation of GDH enabling consistent type determination, stable high-order axis localization, and a texture-aware extension that models appearance-induced reductions in rotational order while preserving axis orientations.
	\item A scalable and annotation-free source of rotational symmetry priors for unseen-object pose estimation. Training FoundationPose with these priors on a one-million-image corpus yields up to a 0.9\% gain on BOP, highlighting the value of symmetry-aware augmentation for foundation-level models.
	\end{mylist}
\section{Related Work}
\label{sec:Related Work}

Rotational symmetry analysis in 3D vision comprises two fundamental tasks: symmetry-type identification\cite{sym-bib1,sym-bib2,sym-bib3,sym-bib4,sym-bib9}, which infers the underlying rotational structure from data, and symmetry-axis localization\cite{sym-bib5,sym-bib6,sym-bib7,sym-bib8,kasal}, which assumes the type is known and estimates its corresponding axes. Existing approaches addressing one or both tasks vary mainly along two design dimensions: (i) reliance on geometric priors such as centroids, keypoints, or object frames, and (ii) coverage of the eight canonical rotational symmetry types. These dimensions strongly influence robustness and generality.

\subsection{Symmetry Type Identification}

Symmetry type detection identifies whether an object exhibits rotational symmetry and, if so, infers its specific type and order. Existing approaches fall into two categories: reference-based and reference-free. Reference-based methods \cite{sym-bib1,sym-bib2} rely on geometric anchors such as centroids, midpoints, or principal axes. They perform well under controlled conditions but are highly sensitive to surface noise, occlusion, and misalignment, which limits their applicability in real-world 6D pose estimation. Reference-free methods \cite{sym-bib3,sym-bib4,sym-bib9} infer symmetry directly from surface geometry through optimization, spectral analysis, or voting. They achieve greater robustness by avoiding predefined anchors but still lack generality across all canonical rotational types. For example, Hruda et al.\cite{sym-bib4} proposed a reflection-based framework that fails on continuously symmetric objects such as cylinders or cones, while Rajkumar et al.\cite{sym-bib3} presented a hybrid approach limited to continuous and second-order discrete conical symmetries.

\subsection{Symmetry Axis Localization}
Most existing methods address only limited cases\cite{sym-bib5,sym-bib6,sym-bib7,sym-bib8,sym-bib10} and are typically designed for specific scenarios. For example, Sipiran\cite{sym-bib6,sym-bib7} proposed approaches for cultural heritage restoration that focuses on continuous types. These methods, however, cannot accommodate texture-induced symmetries. Although Wang et al.\cite{kasal} introduced a method capable of handling all symmetry types, including texture-related cases, it still requires manual specification of symmetry type and order. This dependency hinders full automation and leads to annotation ambiguity, particularly for objects with high-order or composite symmetries that even experts may find difficult to classify accurately.

\subsection{Rotational Symmetry in Pose Estimation}
In 6D pose estimation, existing methods for handling rotationally symmetric objects fall into two categories. The first does not use explicit symmetry priors\cite{bib4,bib5,bib43,epos} but incorporates architectural or learning designs that implicitly capture latent symmetry cues. Examples include EPOS\cite{epos} and GDR-Net\cite{bib43}. The second\cite{bib1,bib2,bib3,es6d,Richter,bib10,hcce} explicitly leverages known symmetry priors to resolve pose ambiguities and improve accuracy. Methods such as HccePose (BF)\cite{hcce} and ZebraPose\cite{bib10} use predefined symmetry matrices to transform ground-truth rotations, selecting the variant closest to the identity as the training target. This strategy reduces symmetry-induced ambiguity and leads to more stable training and higher accuracy than prior-free methods. Symmetry priors also underpin evaluation protocols. Metrics such as  ADD-S\cite{ADDS}, MSSD\cite{MSSD}, and MSPD\cite{MSSD} define pose equivalence based on symmetry-aware criteria; without them, physically identical yet numerically distinct poses would be misjudged. Hence, symmetry priors are essential not only for accurate estimation but also for consistent evaluation.

Despite their importance, existing symmetry annotations remain largely manual or require predefined types or orders, limiting scalability. These limitations motivate the development of a fully automatic framework for robust object symmetry analysis without human intervention. To this end, KASALv2 is introduced, enabling automatic detection of symmetry types and precise axis localization across all eight canonical rotational types, thereby providing reliable symmetry priors for downstream pose estimation.
\section{Methodology}
Our framework performs reference-free rotational symmetry analysis for 3D objects. Central to the method is the Geometric Degeneration Hierarchy (GDH), a computable structure encoding all canonical rotational symmetry families and the degeneration conditions between them. Guided by GDH, the method localizes the dominant high-order axis, estimates its order, determines the symmetry type, and reconstructs the full canonical axis configuration. A texture-aware extension further models appearance-induced order reductions while preserving geometric axis orientations. The overall workflow is shown in Fig.~\ref{fig2:pipline}.

This section follows a theory-to-operation progression. Sec.~\ref{3.1} introduces GDH, which defines the structural relations and degeneration pathways among all rotational symmetry families. Sec.~\ref{3.2} and Sec.~\ref{3.3} then apply these relations to localize the dominant axis, infer its order, classify the symmetry type, and recover all canonical axes. Sec.~\ref{3.4} extends the framework to appearance-aware settings using texture-induced order reductions.

\subsection{Geometric Degeneration Hierarchy (GDH) for 3D Rotational Symmetry}
\label{3.1}
Rotational symmetry in three-dimensional space corresponds to the orientation-preserving subgroups of \(\mathrm{SO}(3)\), including the continuous types \(C_{\infty}\) and \(D_{\infty}\), the finite cyclic and dihedral groups \(C_n\) and \(D_n\), and the rotational groups of the Platonic solids \((A_4, S_4, A_5)\)~\cite{geo1,geo2,geo3,geo4,geo5,geo6,geo7}. These families cover all canonical rotational symmetries considered in this work, with classical subgroup embeddings such as \(A_4 \subset S_4\) and \(A_4 \subset A_5\)~\cite{geo8,geo9}. Building on this foundation, we introduce the Geometric Degeneration Hierarchy (GDH), a geometry-driven and algebraically complete hierarchy that reorganizes these $SO(3)$ subgroups into an operational form. Rather than a purely conceptual taxonomy, GDH encodes how axis multiplicity, rotational order, and characteristic inter-axis relations evolve under geometric degenerations, providing computable constraints that restrict feasible symmetry families and support type-specific axis inference in our fully automatic framework.

At the top of this hierarchy lies the maximally symmetric sphere, where every direction is a valid rotational axis. From this state, GDH considers two geometric degeneration lineages: the dihedral–cyclic family, characterized by reductions from \(C_{\infty}\) and \(D_{\infty}\) to their finite counterparts, and the Platonic family, representing discrete reductions of multi-axis structures such as \(A_5\), \(S_4\), and \(A_4\).

The dihedral–cyclic lineage begins with the cylindrical form \(D_{\infty}\), characterized by continuous rotation about a single principal axis. From this state, symmetry may evolve along two geometric reduction paths. The first is the discretization of continuous rotation,
\begin{equation}
	D_{\infty} \Rightarrow D_{n},\qquad C_{\infty} \Rightarrow C_{n},\qquad n \in \mathbb{N},\; n \ge 2,
\end{equation}
producing finite \(n\)-fold rotational structures with or without orthogonal two-fold axes. This discretization relation allows the dominant axis order to restrict the feasible discrete families during type inference.

The second path corresponds to the loss of orthogonal axes,
\begin{equation}
	D_{\infty} \Rightarrow C_{\infty},\qquad D_{n} \Rightarrow C_{n},
\end{equation}
yielding single-axis rotational configurations.

In parallel, the Platonic lineage models the geometric reduction of multiple high-order axes present in fully symmetric structures. This leads to the icosahedral \((A_{5})\), octahedral \((S_{4})\), and tetrahedral \((A_{4})\) symmetries, which exhibit progressively fewer and lower-order rotational axes:
\begin{equation}
	A_{5} \Rightarrow A_{4},\qquad S_{4} \Rightarrow A_{4}.
\end{equation}

Each downward transition reflects a geometric simplification in both the number and orientation of rotational axes. Together, the dihedral–cyclic and Platonic lineages form the Geometric Degeneration Hierarchy (GDH), a geometry-based organizational framework encompassing all eight canonical types of 3D rotational symmetry.

Within this hierarchy, the localized dominant high-order axis provides the primary constraint that narrows the feasible subtype of symmetry types, while its estimated rotational order further restricts the set of candidates. The identification of a secondary axis that satisfies the corresponding geometric relations is then used to verify the specific type. Once the type is determined, its intrinsic axis relations uniquely specify the remaining canonical axes, enabling complete reconstruction of the object's rotational structure. Through these components, GDH supplies a structured set of geometric constraints that make the subsequent classification and axis reconstruction steps operationally tractable within our fully automatic framework.

\subsection{High-order Axis Localization and Order Identification}
\label{3.2}
\subsubsection{Dominant Axis Localization}

Guided by the GDH framework introduced in Sec.~\ref{3.1}, this stage localizes the dominant high-order rotational axis, which anchors subsequent order identification and type classification. Unlike reference-based methods that depend on pre-aligned templates or centroids, our approach operates entirely in a reference-free manner by optimizing a global alignment objective directly on the unit sphere.

\textbf{Alignment Principle.}

To locate the true rotational axis without prior knowledge of symmetry order, candidate directions are evaluated using a mixed-order alignment loss that aggregates alignment errors across multiple potential orders:
\begin{equation}
{{L}_{mix}}\left( a \right)=\mathop{\sum }_{n\in N}{{L}_{n}}\left( a \right)~
\end{equation}
\begin{equation}
{{L}_{n}}\left( a \right)=\underset{k=1}{\overset{n-1}{\mathop \sum }}\,Chamfer\left( P,R\left( a,{{\theta }_{k}} \right)P \right)
\end{equation}
where ${{\theta }_{k}}=2\pi k/n$ and $R\left( a,{{\theta }_{k}} \right)$ denotes rotation of the point cloud $P$ around axis $a$ by ${{\theta }_{k}}$ and Chamfer$(·,·)$ denotes the Chamfer distance\cite{chamfer}. 

\begin{figure}[t]
	\centering
	\includegraphics[width=0.85\linewidth]{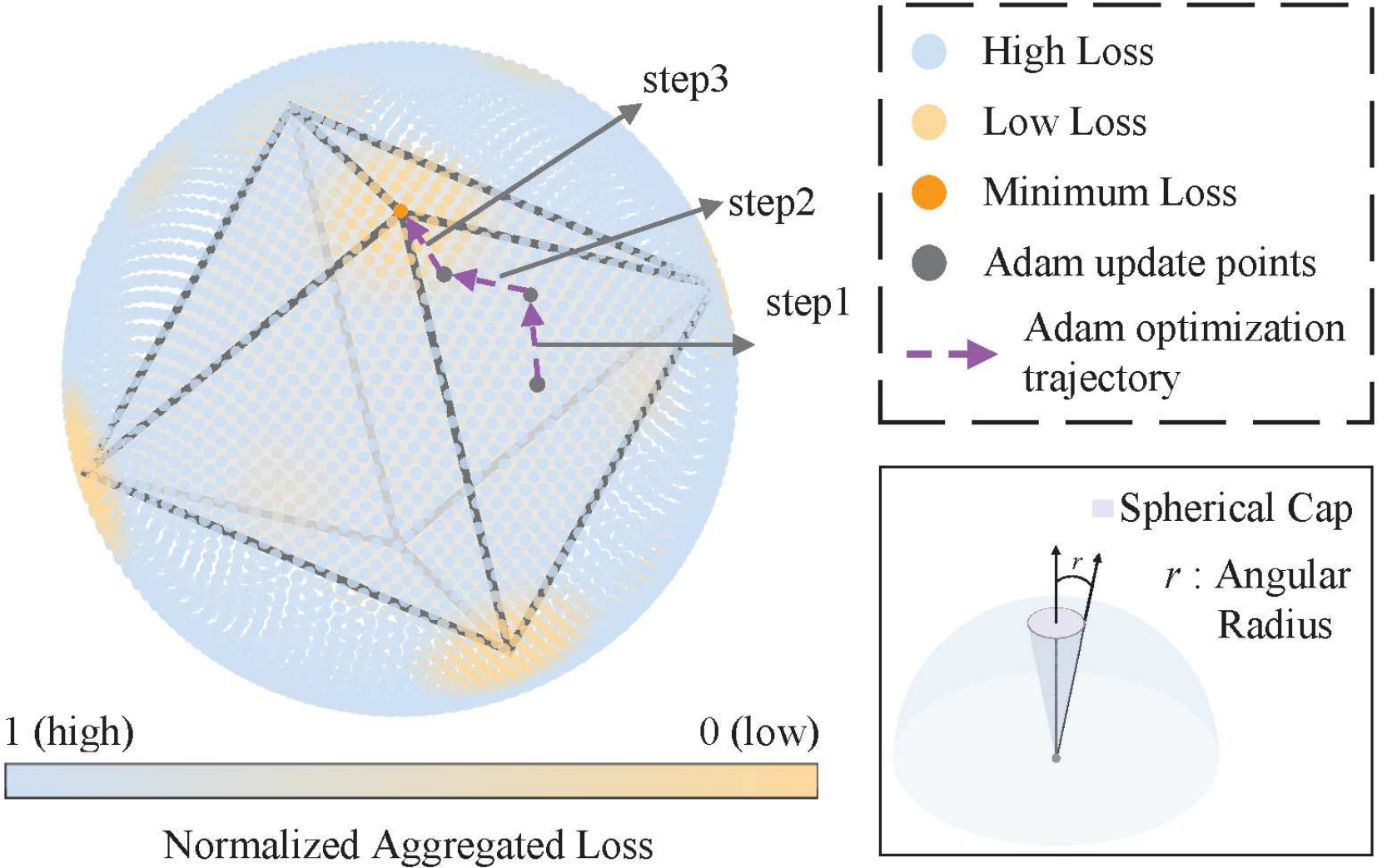}
	\caption{Aggregated alignment-loss distribution on the unit sphere. Low-loss regions (orange) cluster around the true high-order axis, illustrating the self-consistent nature of high-order symmetries.
	}
	\label{fig3:mixloss}
\end{figure}

Because high-order axes inherently satisfy the periodicity of lower-order rotations, their aggregated loss tends to be smaller, forming natural descent basins in the loss landscape (Fig.~\ref{fig3:mixloss}). As a result, optimization is naturally biased toward these dominant high-order directions during search. This self-guiding property enables high-order symmetries to emerge automatically, thereby supporting fully reference-free localization of the dominant axis.

\textbf{Two-Stage Sampling and Optimization.}

To balance global exploration and computational efficiency, axis localization is carried out in two stages: global coarse sampling followed by local fine refinement, with a final gradient-based optimization for convergence stability.

In the first stage, global coarse sampling uniformly explores the unit sphere to identify regions of low mixed-order alignment loss, yielding the top-${{k}_{cand}}$ candidate directions. The second stage performs local fine sampling around each candidate by resampling a spherical cap of angular radius $r$ and re-evaluating top-${{k}_{init}}$ directions under the same loss. This hierarchical search strategy enables efficient localization of promising minima while maintaining broad coverage of the loss landscape. The refined candidates are then optimized using the Adam algorithm to further minimize the aggregated loss and stabilize the alignment result.

Through this progressive sampling-to-optimization pipeline, the framework robustly localizes the dominant high-order rotational axis, which subsequently serves as the geometric anchor for order estimation.

\subsubsection{Rotational Order Identification}

After the dominant axis ${{a}^{\text{*}}}$ is localized, rotational order ${{N}_{\text{est}}}$ is inferred by analyzing the object’s periodic self-alignment under densely sampled rotations. The object is rotated around ${{a}^{\text{*}}}$ through a dense set of uniformly spaced angles  $\phi \in \left[ 0,2\pi  \right)$ (typically 360 samples per full revolution), and the Chamfer distance between the original and rotated shapes is computed at each step to form a rotational self-similarity signal. The dominant frequency of this signal reveals the object’s fundamental rotational periodicity. 

A coarse estimate ${{N}_{fft}}$ is obtained from the dominant frequency component and further refined through direct alignment evaluation over candidate orders $\left\{ {{N}_{fft}},~{{N}_{fft}}\pm 1 \right\}$. The order that minimizes the reconstruction error is selected as the final ${{N}_{\text{est}}}$. Ambiguous or degenerate cases are resolved through hierarchical verification: when ${{N}_{\text{est}}}\ge 3$, the object is considered discretely rotationally symmetric; when ${{N}_{\text{est}}}=2$, a ${{180}^{\circ }}$ rotation test distinguishes genuine two-fold symmetry from reflection-like structures; and when ${{N}_{\text{est}}}\le 1$, a small-angle (${{45}^{\circ }}$) rotation test differentiates continuous from asymmetric shapes.

This procedure provides a robust, reference-free estimation of both the dominant axis and its rotational order, forming a reliable basis for the hierarchical symmetry classification and full-axis reconstruction introduced in Sec.~\ref{3.3}.

\subsection{Hierarchical Type Classification and Full-Axis Reconstruction}
\label{3.3}
Building upon the Geometric Degeneration Hierarchy (GDH) introduced in Section~\ref{3.1}, each symmetry type is characterized by the number and relative orientation of its rotational axes. Once the dominant high-order axis \(a^{*}\) is localized and its order \(N_{\text{est}}\) is inferred, the GDH restricts the hypothesis space to the subset of types consistent with \(N_{\text{est}}\).

The remaining ambiguity concerns the orientation of a secondary high-order axis relative to \(a^{*}\). Under GDH, each candidate symmetry type prescribes a characteristic inter-axis inclination \(\alpha_{F}\) between \(a^{*}\) and a secondary axis \(b^{*}\), which may share the same or a lower rotational order. Because these inclinations are invariant to pose and scale, verifying the existence of a secondary axis at the prescribed \(\alpha_{F}\) serves to identify the correct symmetry type and resolve the remaining ambiguity.

\subsubsection{Secondary-Axis Identification and Hierarchical Type Classification}

The secondary axis ${{b}^{*}}$ is searched along a circular locus on the unit sphere, defined by a fixed inclination ${{\alpha }_{F}}$ from the dominant axis ${{a}^{*}}$, as prescribed by the corresponding GDH type. The optimal direction is obtained by minimizing the alignment loss
\begin{equation}
{{b}^{\text{*}}}=\arg \underset{\,b\in S\left( {{\text{ }\!\!\alpha\!\!\text{ }}_{F}} \right)}{\mathop{\min }}\,{{L}_{\text{Chamfer}}}\left( b;N \right)
\end{equation}
where $N$ denotes the hypothesized order of $b$ (equal to ${{\text{N}}_{\text{est}}}$ for Platonic families or a lower order for prismatic and pyramidal ones).
 
Candidate directions with the lowest losses are preserved as top-$k$ seeds. Each seed undergoes a one-degree-of-freedom gradient refinement along the ring parameter axis, inherently enforcing the inclination constraint. The refinement, performed using the Adam optimizer, minimizes ${{L}_{\text{Chamfer}}}\left( b;N \right)$. Once ${{b}^{\text{*}}}$ is obtained, the symmetry type is determined from the order pair $\left( {{N}_{{{a}^{\text{*}}}}},{{N}_{{{b}^{\text{*}}}}} \right)$ and the theoretical inclination ${{\alpha }_{F}}$ defined in the GDH.

\textbf{Continuous symmetries} $\left( {{C}_{\infty }},{{D}_{\infty }} \right)$ are recognized when no discrete periodicity is found along ${{\text{a}}^{\text{*}}}$. The presence of additional perpendicular two-fold axes distinguishes cylindrical (${{D}_{\infty }}$) from circular (${{C}_{\infty }}$) symmetry, while complete isotropy corresponds to the spherical case.

\textbf{Discrete symmetries} (${{C}_{n}}$, ${{D}_{n}}$) are identified when ${{N}_{est}}\ge 2$. Orthogonal two-fold axes indicate dihedral (${{D}_{n}}$) structures, whereas their absence yields cyclic (${{C}_{n}}$) symmetry.

\textbf{Platonic types} (${{A}_{4}}$, ${{S}_{4}}$, ${{A}_{5}}$) are verified by comparing the measured inter-axis inclination with the theoretical constants ${{70.53}^{\circ }}$ (tetrahedral, ${{A}_{4}}$), ${{90}^{\circ }}$ (octahedral/cubic, ${{{S}}_{4}}$), and ${{63.43}^{\circ }}$ (icosahedral, ${{{A}}_{5}}$).

This geometric classification maps the detected axis configuration to its canonical rotational type, completing symmetry-type determination before the full-axis reconstruction in Sec.~\ref{3.3.2}.
 
\subsubsection{Canonical Axis Reconstruction}
\label{3.3.2}
Once the symmetry type is determined, the GDH provides its canonical axis configuration. With the dominant axis \(a^{*}\) and secondary axis \(b^{*}\) already localized, the remaining axes are recovered by aligning the canonical arrangement associated with the identified type. Each type includes a predefined axis template specifying the relative orientations and multiplicities of its rotational axes. A rigid transformation \(Q\) is estimated to align the template pair \((\hat{a}, \hat{b})\) with the detected pair \((a^{*}, b^{*})\), yielding the full set of axes as
\begin{equation}
	u_i^{\text{final}} = Q\,\widehat{u_i}.
\end{equation}

This alignment transfers type-specific geometric relations to the object, ensuring consistency with its rotational structure. The resulting reconstruction completes full-axis localization in a compact, reference-free form suitable for symmetry reasoning and pose evaluation.

\subsection{Texture-Symmetry Extension for Appearance-Aware Analysis}
\label{3.4}
Real-world objects rarely exhibit perfect geometric regularity. Surface appearance, including texture and color patterns, can break rotational invariance in the image domain. To model this effect, texture is treated as a secondary order reduction applied on top of geometric symmetry. Following the GDH, this reduction alters only the rotational order of each axis while preserving its orientation. The effective orders must lie in the divisor set of the geometric order:
\begin{equation}
	n_{\text{tex}} \in \{\, d \mid d \text{ divides } n_{\text{geo}} \,\} \cup \{1\}.
\end{equation}

Thus, a geometrically six-fold object may appear three- or two-fold under periodic textures, and continuous symmetries such as cylinders can manifest as discrete \(D_n\) patterns when repeated surface structure is present. Given the recovered geometric type and axis configuration, texture refinement is computed analytically by evaluating appearance consistency around each axis, yielding a unified representation that incorporates both geometry- and texture-induced symmetry.
\section{Experiments}
This section presents the experimental setup, ablation studies, and cross-dataset evaluations. We first outline the implementation and evaluation settings, then conduct ablations on the DSRSTO \cite{kasal} dataset. To assess generalization, we further evaluate on GSO\cite{gso} and BOP\cite{VSD}, which include diverse real-world objects with varying symmetry characteristics.

\subsection{Experimental Setup}
\subsubsection{Implementation Details}
All experiments are implemented in Python/PyTorch and executed on an Intel i5-13400F / RTX 3050 GPU and 32 GB RAM. In the high-order axis detection stage, 128 coarse directions are uniformly sampled on the upper hemisphere. The top six are refined within $10^{\circ}$ spherical caps using eight local samples, and the best three undergo Adam optimization (lr=0.04, 5 steps). The mixed-order alignment loss uses the candidate order set $N=\left\{ 3,4,5,6 \right\}$. For secondary-axis localization, Adam is applied with lr=0.05 for 5 steps. The number of top candidates equals twice the estimated rotational order, and each region is refined by eight local samples.

\subsubsection{Adaptive Sampling Mechanism}
\textbf{High-Order Axis Detection.} A single spherical-cap radius $r$ governs all sampling budgets within the valid range $r\in \left[ {{r}_{min}},{{r}_{max}} \right]$, where coverage remains geometrically meaningful. The theoretical lower bound of the coarse-sampling count is derived from the ratio between the sphere surface and the cap area,
\begin{equation}
	N_{min} = \frac{2}{1 - \cos r}
\end{equation}
representing the minimal number of caps required for full coverage. Sampling is limited to the upper hemisphere to avoid antipodal redundancy, and ${{N}_{min}}$ is rounded to the nearest multiple of eight for computational convenience.

The radius $r$ jointly determines the coarse-sampling count ${{N}_{min}}\left( r \right)$, candidate directions ${{k}_{\text{cand}}}\left( r \right)$, local fine samples ${{n}_{\text{local}}}\left( r \right)\ge 4$, and Adam initializations ${{k}_{\text{init}}}\left( r \right)$. Smaller $r$ increases ${{k}_{\text{cand}}}\left( r \right)$ to capture narrow high-order axes, while larger $r$ raises ${{k}_{\text{init}}}\left( r \right)$ for broader exploration. Both are capped at six, consistent with the fact that in 3D finite rotational symmetries the number of maximal-order $(\ge3)$ axes never exceeds six (achieved only by the icosahedral case). This radius-driven formulation balances sampling density and computational efficiency, ensuring stable localization across geometries.

\textbf{Secondary-Axis Localization.} Sampling density scales linearly with the target rotational order, allocating proportionally more candidates and local samples for higher orders to maintain consistent angular resolution.

\subsubsection{Metrics and Datasets}
\label{4.1.3}
To quantitatively evaluate the proposed framework, both classification and localization metrics are employed.

\textbf{Classification Metrics.} To assess symmetry-type and order prediction, we report three accuracies:
\begin{subequations}
	\begin{align}
		acc_t &= \frac{N_{\mathrm{correct\ type}}}{N_{\mathrm{all}}}, \\[4pt]
	acc_o &= \frac{N_{\mathrm{correct\ order\ |\ correct\ type}}}{N_{\mathrm{correct\ type}}}, \\[4pt]
		acc_T &= \frac{N_{\mathrm{correct\ type\ \&\ order}}}{N_{\mathrm{all}}}.
	\end{align}
\end{subequations}
These metrics respectively reflect type accuracy, order-consistent accuracy, and overall correctness.

\textbf{Localization Metric.} Axis localization is evaluated using the Average Distance of Model Points (ADI), which measures the discrepancy between the original model and its reconstructed symmetric instances. Given the predicted symmetry orders, directions, and centers, the rotational symmetry is represented by a discrete rotation set $T_{p}={T_{i}}$. The localization error is
\begin{equation}
	e_{\text{ADI}}=\frac{\sum_{T_{i}\in T_{p}}\text{ADI}(T_{i},I)}{|T_{p}|},
\end{equation}
where $\text{ADI}(T_{i},I)$ denotes the mean pointwise distance between the model and its rotated counterpart.

Misclassified objects yield incorrect canonical axis groups and thus invalid rotation sets, making $e_{\text{ADI}}/d$ meaningless. Accordingly, $e_{\text{ADI}}/d$ is reported only for samples correctly identified in both type and order.

\textbf{Datasets.} The DSRSTO\cite{kasal} dataset includes 38 CAD objects across seven categories, covering all canonical 3D rotational symmetry types with 27 geometrically symmetric and 11 texture-symmetric models. As a manually designed benchmark, it serves to evaluate symmetry classification and axis localization. To assess performance in real scenarios, tests are conducted on common 6D pose benchmarks such as ITODD\cite{itodd}, T-LESS\cite{tless}, LM-O\cite{adi}, HB\cite{hb}, YCB-V\cite{ADDS}, and IC-BIN\cite{icbin}, which contain real-world objects but limited symmetry diversity. The GSO\cite{gso} dataset extends the evaluation to 944 scanned models, among which 438 exhibit rotational symmetry, providing a large-scale testbed for generalization across geometries.
\subsection{Ablation Studies}
Ablation experiments on the DSRSTO dataset were conducted to analyze the influence of core design choices on classification accuracy, axis-localization precision, and computational efficiency. The evaluation considers overall accuracy $acc_T$, normalized geometric alignment error ${{\text{e}}_{\text{ADI}}}/d$ (where $d$ is the object diameter), and average runtime per object. Two factors were examined: the adaptive radius-driven sampling configuration and the composition of the candidate order set.
\begin{figure}[t]
	\centering
	\includegraphics[width=0.90\linewidth]{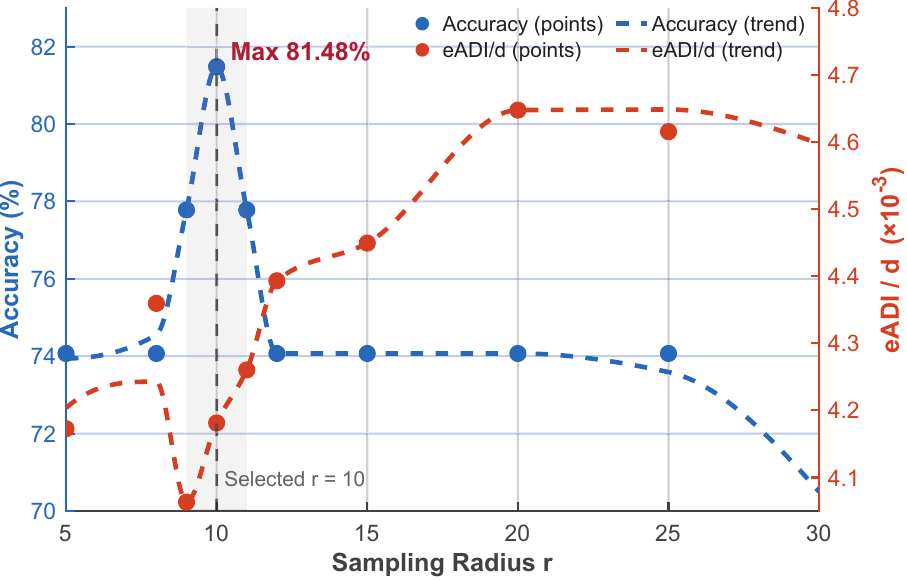}
	\caption{Effect of the spherical-cap radius $r$ on sampling performance.
		Dashed curves indicate fitted trends for total accuracy (blue) and
		normalized alignment error $e_{\text{ADI}}/d$ (orange). The optimal
		configuration appears near $r = 10^{\circ}$.}
	\label{fig1:onecol}
		
	\vspace{4pt} 
	
	\captionof{table}{Ablation experiments on candidate order sets $\mathcal{N}$.}
	\vspace{8pt}
	\label{tab:order_set_ablation}
	\footnotesize
	\begin{tabular}{l|c|c|c}
		\toprule
		Order Set $N$     & $acc_T$ (\%) & $e_{\text{ADI}}/d$ & Time (s) \\
		\midrule
		$\{3,4\}$         & 70.37        & 0.00419            & 1.02     \\
		$\{3,4,5\}$       & 77.78        & 0.00423            & 1.20     \\
		$\{3,5,6\}$       & 77.78        & 0.00411            & 1.69     \\
		$\{3,4,5,6\}$     & 81.48        & 0.00418            & 1.46     \\
		$\{3,4,5,6,7\}$   & 81.48        & 0.00419            & 1.77     \\
		$\{3,4,5,6,7,9\}$ & 81.48        & 0.00429            & 2.40     \\
		$\{2,3,4,5,6\}$   & 77.78        & 0.00659            & 2.28     \\
		\bottomrule
	\end{tabular}
\end{figure}
\subsubsection{Sampling Configuration}
The effect of the spherical-cap radius $r$ was evaluated within $\left[ {{5}^{\circ }},{{30}^{\circ }} \right]$ to assess its impact on coverage, accuracy, and efficiency. As shown in Fig.~\ref{fig1:onecol}, performance forms a single peak near $r={{10}^{\circ }}$. Both total accuracy and normalized alignment error remain stable between ${{9}^{\circ }}$ and ${{11}^{\circ }}$, achieving the optimum at $r={{10}^{\circ }}$ ($acc_T=81.48\%$, ${{e}_{ADI}}/d=4.18\times {{10}^{-3}}$). Smaller radii ($\le {{8}^{\circ }}$) lead to redundant refinements without accuracy gain, while larger ones ($\ge {{12}^{\circ }}$) reduce coverage and precision. Runtime varies slightly ($\approx 1.3-1.8s$ per model). The interval ${{9}^{\circ }}-{{11}^{\circ }}$ thus provides the best trade-off among coverage, accuracy, and efficiency, and $r={{10}^{\circ }}$ is adopted as the default.
\subsubsection{Order-Set Composition}
Table~\ref{tab:order_set_ablation} summarizes the evaluation of different candidate order sets $\mathcal{N}$, obtained by systematically reducing, expanding, and cross-validating their elements, including configurations with and without the 2-fold term.

Reducing $\mathcal{N}$ to $\left\{ 3,4 \right\}$ or $\left\{ 3,4,5 \right\}$ caused clear accuracy drops, indicating that $\left\{ 3,4,5,6 \right\}$ is the minimal effective configuration. Expanding the set to $\left\{ 3,4,5,6,7 \right\}$ or $\left\{ 3,4,5,6,7,9 \right\}$ increased runtime without measurable benefit, while $\left\{ 3,5,6 \right\}$ further confirmed the necessity of the 4-fold term for distinguishing cubic and dihedral symmetries. Including $n=2$ introduced instability, as trivial half-turn alignments dominated optimization and led to high-order shapes, particularly several 3-fold models, being misclassified as 2-fold. (Detailed analyses are provided in the supplementary material.) Overall, $\mathcal{N}=\left\{ 3,4,5,6 \right\}$ provides the best balance among accuracy, stability, and computational efficiency, while adding 2-fold or rare high-order terms introduces misalignment or unnecessary cost.
\begin{table*}[t]
	\centering
	\caption{Comprehensive evaluation of symmetry classification and axis localization across datasets. DSRSTO$_{\mathrm{tex}}$: texture subset.}
	\footnotesize
	\setlength{\tabcolsep}{3.8pt}
	\renewcommand{\arraystretch}{0.9}
	\setlength{\aboverulesep}{1pt}
	\setlength{\belowrulesep}{1pt}
	\setlength{\cmidrulesep}{0pt}
	\label{crosstesult}
	\begin{tabular}{l|cc|ccc|cccccc}
		\toprule
		\multirow{2}[0]{*}{Method} & \multicolumn{2}{c|}{Dataset} & \multicolumn{3}{c|}{Category Statistics} & \multicolumn{5}{c}{Dataset-Level Performance} \\
		& Name  & Object Type & Count & Errors & $acc_T$ & $acc_t$ & $acc_o$ & $acc_T$ & time  & \(e_{\mathrm{ADI}}/d\) (ours / \text{Wang et al.}~\cite{kasal}) \\
		\midrule
		\multicolumn{1}{c|}{\multirow{4}[0]{*}{BOP (original)}} & \multicolumn{1}{c}{\multirow{4}[0]{*}{BOP\cite{VSD}}} & ${C}_{\infty}$   & 7     & 0     & 100.00\% & \multirow{4}[0]{*}{80.00\%} & \multirow{4}[0]{*}{100.00\%} & \multirow{4}[0]{*}{80.00\%} & \multirow{4}[0]{*}{—} & \multirow{4}[0]{*}{—} \\
		&       & ${D}_{\infty}$   & 3     & 0     & 100.00\% &       &       &       &       &  \\
		&       & $C_n$  & 35    & 9     & 74.29\% &       &       &       &       &  \\
		&       & $D_n$  & 5     & 1     & 80.00\% &       &       &       &       &  \\
		\midrule
		\multirow{4}[0]{*}{KASALv2(ours)} & \multirow{4}[0]{*}{BOP\cite{VSD}} & ${C}_{\infty}$   & 7     & 0     & 100.00\% & \multirow{4}[0]{*}{92.00\%} & \multirow{4}[0]{*}{91.30\%} & \multirow{4}[0]{*}{84.00\%} & \multirow{4}[0]{*}{1.61} & \multirow{4}[0]{*}{0.00256/0.00290} \\
		&       & ${D}_{\infty}$   & 3     & 0     & 100.00\% &       &       &       &       &  \\
		&       & $C_n$  & 35    & 8     & 77.14\% &       &       &       &       &  \\
		&       & $D_n$  & 5     & 0     & 100.00\% &       &       &       &       &  \\
		\midrule
		\multirow{5}[0]{*}{KASALv2(ours)} & \multirow{5}[0]{*}{GSO\cite{gso}} & ${C}_{\infty}$   & 118   & 3     & 97.46\% & \multirow{5}[0]{*}{96.58\%} & \multirow{5}[0]{*}{98.10\%} & \multirow{5}[0]{*}{94.75\%} & \multirow{5}[0]{*}{1.37} & \multirow{5}[0]{*}{0.00262/0.00256} \\
		&       & ${D}_{\infty}$   & 11    & 0     & 100.00\% &       &       &       &       &  \\
		&       & $C_n$  & 85    & 14    & 83.53\% &       &       &       &       &  \\
		&       & $D_n$  & 222   & 6     & 97.30\% &       &       &       &       &  \\
		&       & $S_4$  & 2     & 0     & 100.00\% &       &       &       &       &  \\
		\midrule
		\multirow{7}[0]{*}{KASALv2(ours)} & \multirow{7}[0]{*}{DSRSTO\cite{kasal}} & ${C}_{\infty}$   & 1     & 0     & 100.00\% & \multirow{7}[0]{*}{81.48\%} & \multirow{7}[0]{*}{100.00\%} & \multirow{7}[0]{*}{81.48\%} & \multirow{7}[0]{*}{1.46} & \multirow{7}[0]{*}{0.00212/0.00172} \\
		&       & ${D}_{\infty}$   & 1     & 0     & 100.00\% &       &       &       &       &  \\
		&       & $C_n$  & 7     & 5     & 28.57\% &       &       &       &       &  \\
		&       & $D_n$  & 12    & 0     & 100.00\% &       &       &       &       &  \\
		&       & $A_4$  & 1     & 0     & 100.00\% &       &       &       &       &  \\
		&       & $S_4$  & 3     & 0     & 100.00\% &       &       &       &       &  \\
		&       & $A_5$  & 2     & 0     & 100.00\% &       &       &       &       &  \\
		\midrule
		\addlinespace[2pt]
		KASALv2(ours) & $\text{DSRSTO}_\text{tex}$\cite{kasal} & $D_n$  & 11    & 3     & 72.73\% & 72.32\% & 87.50\%  & 72.32\% & 1.16  & 0.00195/0.00181 \\
		\bottomrule
	\end{tabular}
\end{table*}
\subsection{Cross-Dataset Evaluation}
Cross-dataset evaluations were conducted on DSRSTO, BOP, and GSO. Table~\ref{crosstesult} reports the metrics defined in Sec.~\ref{4.1.3}, the average runtime per object, and localization accuracy compared with the reference method \cite{kasal}.

On DSRSTO, the method achieved 81.48\% accuracy with \(e_{\mathrm{ADI}}/d=0.00212\). On BOP, under our stricter re-annotation of 50 symmetric instances, it reached 84.00\% accuracy, improving over 80.00\% under the original labels and reducing \(e_{\mathrm{ADI}}/d\) from 0.00290 to 0.00256. On GSO, it obtained 94.75\% accuracy with \(e_{\mathrm{ADI}}/d=0.00262\). The 72.32\% accuracy on DSRSTO$_{\mathrm{tex}}$ further confirms applicability to texture-level symmetries. The lower DSRSTO performance mainly stems from challenging high-order and visually ambiguous \(C_n\) cases; consistently, \(C_n\) is also the lowest-accuracy category in BOP. Details of the original BOP annotations, our re-annotation, and the full error analysis are provided in the supplementary material.

\subsection{Effect of Symmetry Priors on Pose Estimation}
\label{4.4}
To assess the downstream value of automatically estimated symmetry priors, we train two identical FoundationPose~\cite{FP} models on a synthetic GSO training set of about one million rendered images, differing only in the use of KASALv2-derived rotational-symmetry priors. The prior is injected only at the label level via ZebraPose's symmetry-aware label canonicalization module, which maps each GT pose to a canonical representative in its symmetry equivalence class for both discrete and continuous symmetries. All other settings remain unchanged.

Table~\ref{tab:foundationpose_kasalv2} reports a retrained multi-checkpoint evaluation on five BOP datasets. Averaged over five checkpoints from the same training run, the results improve on all five datasets, with the largest mean gain of 0.9 points on IC-BIN.

\begin{table}[t]
	\centering
	\small
	{
		\setlength{\tabcolsep}{9pt}
		\renewcommand{\arraystretch}{0.95}
		\footnotesize
		\caption{FoundationPose with and without KASALv2 symmetry priors on five BOP datasets. Results are reported as mean (range) over five checkpoints from the same training run (seed 0).}
		\label{tab:foundationpose_kasalv2}
		\begin{tabular}{lccc}
			\toprule
			\textbf{Dataset} &
			\makecell{\textbf{w/ KASALv2}\\\textbf{mean (range)}} &
			\makecell{\textbf{w/o KASALv2}\\\textbf{mean (range)}} &
			\makecell{\textbf{Diff.}} \\
			\midrule
			LM-O   & 71.9 (71.7--72.0) & 71.6 (71.5--71.8) & +0.3 \\
			IC-BIN & 64.7 (64.4--65.3) & 63.8 (63.2--64.4) & +0.9 \\
			YCB-V  & 87.7 (87.6--87.9) & 87.6 (87.5--87.8) & +0.1 \\
			ITODD  & 65.8 (65.1--66.4) & 65.0 (64.9--65.2) & +0.8 \\
			T-LESS & 59.3 (58.6--60.1) & 58.8 (58.3--59.3) & +0.5 \\
			\bottomrule
		\end{tabular}
	}
\end{table}
\section{Conclusion}
This work presents a unified and fully automatic framework for 3D rotational symmetry analysis, jointly addressing symmetry-type classification, rotational-order identification, and axis localization across all eight canonical types. The Geometric Degeneration Hierarchy (GDH), together with an integrated inference pipeline, enables complete and reference-free symmetry reasoning for real-world 3D objects. Extensive experiments show strong accuracy and generalization, reaching 94.75\% on GSO, and the symmetry priors produced by KASALv2 improve FoundationPose accuracy by up to 0.9\% across five BOP datasets. Despite these results, characteristic failure modes remain, including the underestimation of high rotational orders and sensitivity to local geometric irregularities. Future work will explore more frequency-sensitive order estimation and improved robustness to locally perturbed structures.

\clearpage
\section{Acknowledgments}
This work was supported by the National Natural Science Foundation of China under Grant No.~52375487 and the Special Fund of Jiangsu Province for Key Research and Development under Grant No.~BE2023041.

The authors also thank the reviewers for their thoughtful and constructive comments, which have helped improve this paper. Due to space limitations, several complementary analyses and extended discussions could not be included in the present conference version and will be reported in a future journal extension, including posterior verification for continuous symmetries and secondary two-fold axes, robustness under geometric perturbations, additional optimizer and runtime analyses, large-scale rotational-symmetry annotation on Objaverse, and broader downstream evaluation studies.
{
    \small
    
    \bibliographystyle{unsrt}
    \bibliography{main}
}

\end{document}